# Zero-Shot Motor Health Monitoring by Blind Domain Transition

Serkan Kiranyaz, Ozer Can Devecioglu, Amir Alhams, Sadok Sassi, Turker Ince, Osama Abdeljaber, Onur Avci, and Moncef Gabbouj, *Fellow, IEEE*

*Abstract*— Continuous long-term monitoring of motor health is crucial for the early detection of abnormalities such as bearing faults (up to 51% of motor failures are attributed to bearing faults). Despite numerous methodologies proposed for bearing fault detection, most of them require normal (healthy) and abnormal (faulty) data for training. Even with the recent deep learning (DL) methodologies trained on the labeled data from the same machine, the classification accuracy significantly deteriorates when one or few conditions are altered, e.g., a different speed or load, or for different fault types/severities with sensors placed in different locations. Furthermore, their performance suffers significantly or may entirely fail when they are tested on *another* machine with entirely different healthy and faulty signal patterns. To address this need, in this pilot study, we propose a zero-shot bearing fault detection method that can detect any fault on a new (target) machine regardless of the working conditions, sensor parameters, or fault characteristics. To accomplish this objective, a 1D Operational Generative Adversarial Network (Op-GAN) first characterizes the transition between normal and fault vibration signals of (a) source machine(s) under various conditions, sensor parameters, and fault types. Then for a target machine, the potential faulty signals can be generated, and over its *actual* healthy and *synthesized* faulty signals, a compact, and lightweight 1D Self-ONN fault detector can then be trained to detect the real faulty condition in real time whenever it occurs. To validate the proposed approach, a new benchmark dataset is created using two different motors working under different conditions and sensor locations. Experimental results demonstrate that this novel approach can accurately detect any bearing fault achieving an average recall rate of around 89% and 95% on two target machines regardless of its type, severity, and location.

## I. INTRODUCTION

ELECTRICAL machines with rotating elements are crucial for various industrial applications and transportation systems. Therefore, detecting the fault on a new machine at the moment it occurs is critical to prevent hazardous accidents and damages. According to the IEEE Industry Application Society (IEEE-IAS) [1]-[3] and the Japan Electrical Manufacturers' Association (JEMA) [4], around 40% of all machine faults occur due to bearing faults. Another study in [5] finds that up to 51% of motor failures are attributed to bearing faults. Numerous methodologies have been proposed for bearing fault detection and diagnosis such as model-based methods [6]-[8], traditional signal-processing approaches, [9]-[15] and more recent deep learning (DL) based methods, [17]-[26]. A recent survey article [27] dedicated to bearing fault diagnosis covers more than 180 papers, around 80 of which employed some type of DL paradigms. During the last few years, the number of articles grows exponentially, indicating a booming interest in employing DL methods for bearing fault diagnostics. Those studies mainly show that with proper training, DL methods can achieve high fault detection accuracies even under noisy environments and certain versatile conditions. However, most DL models such as [24]-[26] have a high computational complexity in common which prevents their usage in low-power computing environments in real-time. Additionally, they require a substantial amount of labeled data for training which may not be feasible for a practical application. To address this, the landmark study on the 1D Convolutional Neural Networks [28] and later Self-Organized Operational Neural Networks [29] have recently been proposed and achieved *state-of-the-art* detection accuracies. Most of the aforementioned methods were trained and evaluated in one of the three benchmark datasets, the Case Western Reserve University (CWRU) bearing dataset [30], the Paderborn University bearing dataset [31], and the Intelligent Maintenance Systems (IMS) dataset [32]. To date, most of the proposed methods in the literature on bearing fault identification with ML or DL algorithms employ the CWRU dataset due to its simplicity and popularity.

The aforementioned fault detection methods assume the availability of both healthy and faulty data of the machine for which the fault detection will be performed. The assumption of

S. Kiranyaz is with the Electrical Engineering Department, Qatar University, Doha, Qatar (e-mail: mkiranyaz@qu.edu.qa). A. Alhams, and S. Sassi are with the Mechanical Engineering Department, Qatar University, Doha, Qatar (e-mail: aa1702913@qu.edu.qa; sadok.sassi@qu.edu.qa )

T. Ince, is with the Electrical and Electronics Engineering Department, Izmir University of Economics, Izmir, Turkey (email:turker.ince@ieu.edu.tr).

O. Devecioglu and M. Gabbouj are with the Department of Computing Science, Tampere University, Tampere, Finland (e-mail: moncef.gabbouj@tuni.fi , ozer.devecioglu@tuni.fi ).

O. Abdeljaber is with the Department of Building Technology, Linnaeus University, Växjö, Sweden (email: osama.abdeljaber@lnu.se)

O. Avci is with the Department of Civil and Environmental Engineering, West Virginia University, Morgantown, WV, USA (email: onur.avci@mail.wvu.edu).

having the faulty data on a target machine and especially for all of its working conditions (e.g., motor type, speed, load, etc.) and all sensory locations is obviously not reasonable especially when a new (healthy) machine just becomes operational. With the absence of *faulty* data, those aforementioned *supervised* methods cannot be used to monitor the health of a new machine. Even if faulty data is available, variations in the speed and/or load, different sensor locations, and different fault types or severities may cause significant degradations in the detection performance [27]. This is evident in Figure 1 where eight sample vibration signals acquired from two different (e.g., source and target) machines under different working conditions are shown. The dashed arrows indicate the similar signal patterns between *different* types (healthy *vs* faulty) that may occur not only at different machines but even at the same machine. Moreover, entirely different signal patterns in time and frequency domains may also belong to the same (e.g., healthy) condition.

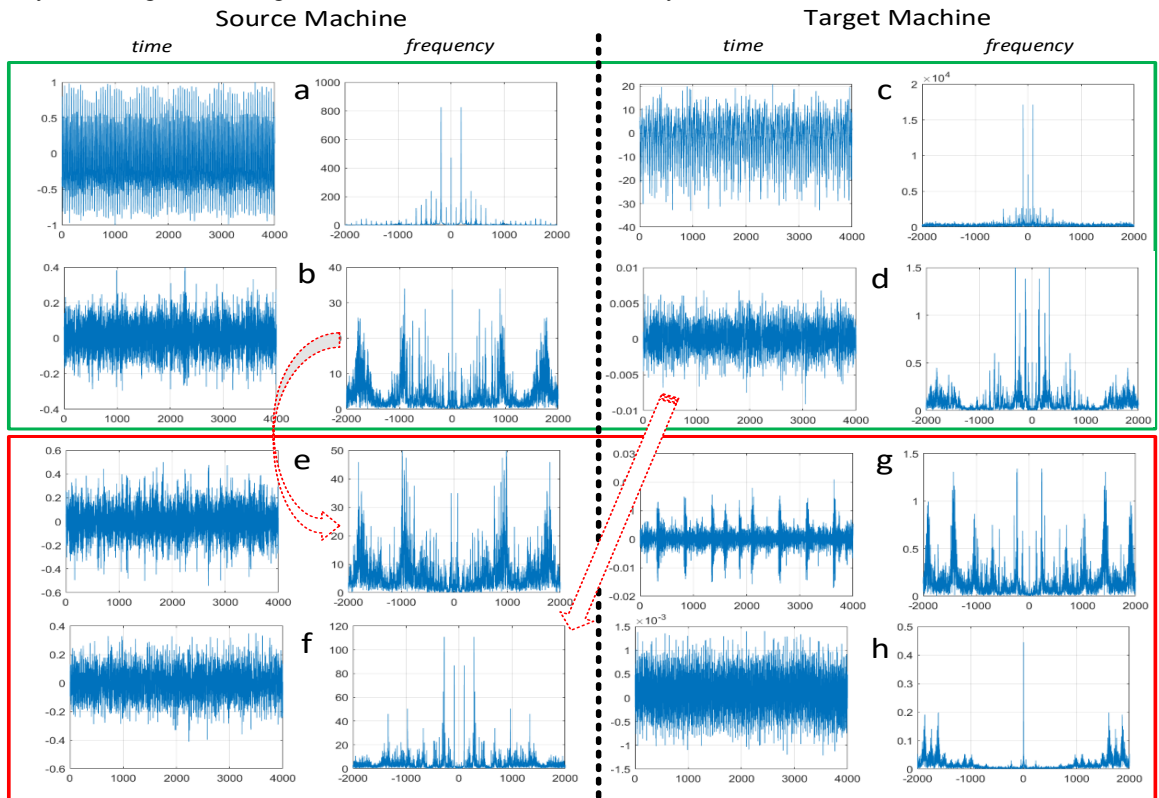

**Figure 1:** Sample vibration signals (time and frequency) from source (left) and target (right) machines. Healthy (a-d) and faulty (e-h) signals are encircled in green and red colors, respectively. Dashed arrows show two similar temporal/spectral patterns from different bearing conditions (healthy and faulty).

In order to address this, domain adaptation (DA) methods have been proposed to transfer knowledge from the *source* domain data to *target* domain data drawn from a different but assuming a *related* distribution. Particularly for the fault detection problem, some recent approaches have been developed to accomplish DA [33]-[44]. The main idea is to adapt (transform) the target domain features to the source domain usually with a regressor network so that the deep classifier trained over the source domain features can generalize sufficiently well on the *transformed* target-domain features. To accomplish this, it is a common practice that another (discriminator) classifier is also trained to distinguish between the source and target features and thus to produce such target features that can be assumed to be *invariant* from the source domain features. There are some other tricks to accomplish this so that the source classifier can be used directly on the transformed features of the target domain. However, DA methods were mostly evaluated on the *same* machine while changing usually one or few working conditions (e.g., only speed or load), sensor parameters (e.g., location), or fault type or severity. Even on the same machine, the assumption of a "related" distribution occurring between the source and target domains (conditions) may fail, and hence, low detection accuracies can be obtained. For instance, a recent *state-of-the-art* method proposed in [44] has reported fault detection accuracies of less than 75% in some cases although the experiments were carried out using the same database (on the same machine) but at a different working condition using the deep and complex networks mostly trained with data augmentation. Obviously, when the source and target domains are on different machines with entirely different working conditions, motors, sensor locations, and faults, the assumption of having a "related distribution" may no longer hold, and hence a significant performance deterioration will occur. The same study [44] also reported that the detection performance drastically suffers for all DA methods when using compact (shallow) networks as the classifier, e.g., accuracies around 50% or even less were obtained even when using the most recent DA methods.

In this study, we propose a zero-shot bearing fault detection method that can detect the actual (unseen) faults on a new (target) machine regardless of its working conditions, sensory setup, or fault severities. To accomplish this objective, we propose blind domain transition using 1D Operational Generative Adversarial Networks (Op-GANs) that learn and thus effectively characterize the natural transition from the healthy to the faulty state of a bearing in the signal domain using the labeled vibration data of one or more source machines. Then for any (new) target machine, the potential faulty signals can be generated directly from its healthy signals under its working conditions. Using the actual healthy and synthesized faulty signals, a compact and shallow classifier, e.g., a lightweight 1D Self-ONN model [45]-[56], can then be trained and used thereafter to detect any actual faulty signal in real time - if and when it occurs. We can summarize the novel and significant contributions of this study as follows:

1- This is a pioneer study where the domain transition to synthesize the unseen fault data is addressed as a "blind" approach thus avoiding any prior assumption between the source and target domains.

2- This is also the first study where 1D Op-GANs are proposed for blind domain transition.

3- This study demonstrates for the first time that the proposed blind domain transition can generate such highly relevant fault signals that hinder the need for traditional DA requirements such as i) training a classifier in the *source* domain and transforming the features extracted in the target domain to source domain in such a way that the source classifier can be used, ii) training a deep and complex network as a classifier to boost the learning performance using several tricks such as data augmentation, dropout, etc.

4- This study further shows that a compact classifier (1D Self-ONN) can be trained directly on the *target* domain to achieve the zero-shot detection of any *actual* faulty signal in real-time with an exceptional performance.

5- Finally, to validate our objectives, the proposed novel approach is evaluated over a new benchmark dataset that contains the vibration data of the two different electric machines under various working conditions, sensor locations, and faulty severities. As a result, a highly varying and diverse set of vibration data (e.g., see some samples in Figure 1) with a total duration of 13.5 hours was acquired that can now serve as the benchmarking testbed for a reliable and full-scale evaluation of zero-shot fault detection methods. The benchmark dataset, our results, and the optimized PyTorch [57] implementation of the proposed approach are now publicly shared[1] with the research community [58].

The rest of the paper is organized as follows: The proposed zero-shot fault detection approach will be presented in Section II. In Section III, the new benchmark vibration dataset will first be introduced and then a detailed set of experimental results will be presented while the performance of the proposed technique is rigorously evaluated. Finally, Section IV concludes the paper and suggests topics for future research.

## II. THE PROPOSED APPROACH

In this section, we first briefly summarize 1D Self-ONNs and their main properties. Then we introduce the proposed 1-D Op-GANs for blind domain transition and the synthesis of the potential faulty data for the target machine. Finally, we shall present the proposed fault detection with a compact classifier trained directly on the target domain.

### A. 1D Self-Organized Operational Neural Networks

Self-ONNs[1] [45]-[56] are the recent variants of ONNs [59]-[61] which were originally derived from Generalized Operational Perceptrons (GOPs) [62]-[67] in the same way conventional CNNs were derived from MLPs. Therefore, instead of the sole linear neuron model of CNNs, Self-ONNs contain *generative* neurons which can generate any arbitrary and nonlinear nodal function, $\Psi$, for each kernel element of each connection. It has been demonstrated in many studies [45]-[56] that both 1D and 2D Self-ONNs can outperform even deep and more complex CNNs with a significant performance gap thanks to the ability to optimize the nodal operator functions during training. To reflect this superiority in blind domain transition, 1D (operational) Cycle-GANs have recently been used for ECG restoration in [56] where the convolutional layers/neurons of native 1D Cycle-GANs (both the generator and discriminator) are replaced by the operational layers with generative neurons of the Self-ONNs.

In a 1D Self-ONN, each kernel element of any generative neuron can perform any nonlinear transformation, $\psi$, the function which can be expressed by the Taylor-series near the origin ($a = 0$),

$$\psi(x) = \sum_{n=0}^{\infty} \frac{\psi^{(n)}(0)}{n!} x^n \qquad (1)$$

The $Q^{th}$ order truncated approximation, formally known as the Taylor polynomial, takes the form of the following finite summation:

$$\psi(x)^{(Q)} = \sum_{n=0}^{Q} \frac{\psi^{(n)}(0)}{n!} x^n \qquad (2)$$

The above formulation can approximate any function $\psi(x)$ near 0. When the activation function bounds the neuron's input feature maps in the vicinity of 0 (e.g., *tanh*), the formulation in (2) can be exploited to form a composite nodal operator where the power coefficients, $\frac{\psi^{(n)}(0)}{n!}$, can be the parameters of the network learned during training.

It was shown in [49], [50], and [52] that the 1D nodal operator of the $k^{th}$ generative neuron in the $l^{th}$ layer can take the following general form:

$$\widetilde{\psi_k^l}\left(w_{ik}^{l(Q)}(r), y_i^{l-1}(m+r)\right) = \sum_{q=1}^{Q} w_{ik}^{l(Q)}(r,q) \left(y_i^{l-1}(m+r)\right)^q \qquad (3)$$

---

[1] The optimized PyTorch source codes, benchmark dataset, and our results are now publicly shared with the research community in https://github.com/OzerCanDevecioglu/Zero-Shot-Bearing-Fault-Detection-by-Blind-Domain-Transition.

Let $x_{ik}^l \in \mathbb{R}^M$ be the contribution of the $i^{th}$ neuron at the $(l-1)^{th}$ layer to the input map of the $l^{th}$ layer. Therefore, it can be expressed as,

$$\widetilde{x_{ik}^l}(m) = \sum_{r=0}^{K-1} \sum_{q=1}^{Q} w_{ik}^{l(Q)}(r,q) \left(y_i^{l-1}(m+r)\right)^q \quad (4)$$

where $y_i^{l-1} \in \mathbb{R}^M$ is the output map of the $i^{th}$ neuron at the $(l-1)^{th}$ layer, $w_{ik}^{l(Q)}$ is a learnable kernel of the network, which is a $K \times Q$ matrix, i.e., $w_{ik}^{l(Q)} \in \mathbb{R}^{K \times Q}$, formed as, $w_{ik}^{l(Q)}(r) = [w_{ik}^{l(Q)}(r,1), w_{ik}^{l(Q)}(r,2), \dots, w_{ik}^{l(Q)}(Q)]$. By the commutativity of the summation operations in (4), one can alternatively write:

$$\widetilde{x_{ik}^l}(m) = \sum_{q=1}^{Q} \sum_{r=0}^{K-1} w_{ik}^{l(Q)}(r, q-1) y_i^{l-1}(m+r)^q \quad (5)$$

One can simplify this as follows:

$$\widetilde{x_{ik}^l} = \sum_{q=1}^{Q} Conv1D\left(w_{ik}^{l(Q)}, \left(y_i^{l-1}\right)^q\right) \quad (6)$$

Hence, the formulation can be accomplished by applying Q 1D convolution operations. Finally, the output of this neuron can be formulated as follows:

$$x_k^l = b_k^l + \sum_{i=0}^{N_{l-1}} x_{ik}^l \quad (7)$$

where $b_k^l$ is the bias associated with this neuron. The $0^{th}$ order term, $q = 0$, the DC bias, is ignored as its additive effect can be compensated by the learnable bias parameter of the neuron. With the $Q = 1$ setting, a *generative* neuron reduces back to a convolutional neuron.

The raw-vectorized formulations of the forward propagation, and detailed formulations of the Back-Propagation (BP) training in raw-vectorized form can be found in [46], [47], and [49].

### B. Blind Domain Transition by 1D Operational GANs

The proposed approach for blind domain transition has the same fundamental motivation of the work in [68] where a personalized early arrhythmia detection method was proposed for a healthy person using his/her ECG signal. In [68], the objective was to detect an unseen arrhythmic ECG signal during regular ECG monitoring. The authors proposed a simple arrhythmia ECG beat syntheses (ABS) method using linear filters, the parameters of which are optimized by the Least-Squares method. Since a normal (healthy) ECG signal has a well-known QRS pattern, such simple linear filters can indeed capture the transition characteristics from the healthy to arrhythmic state in the signal domain accurately and efficiently. However, detecting any unseen faulty on a different (target) machine is significantly more challenging, especially when using vibration signals with no specific pattern or spectrum. The problem intensifies especially due to the variations in the working conditions, sensor parameters, fault types, and severities.

To capture (learn) the transition from the healthy to faulty state in the signal domain accurately, in this study we, therefore, implement 1D Op-GANs where both generator and discriminator networks are 1D Self-ONNs. This gives us the capability of blind characterization independent from the aforementioned variations. The ultimate objective is to capture the transition from the healthy to faulty states in the generator of the GAN and thus, one can utilize this capability to synthesize the potential faulty signals directly from the healthy signals of the new (target) machine. This will enable us to train a simple classifier to distinguish the healthy and faulty signals directly in the target domain without the need for any feature engineering or adaptation.

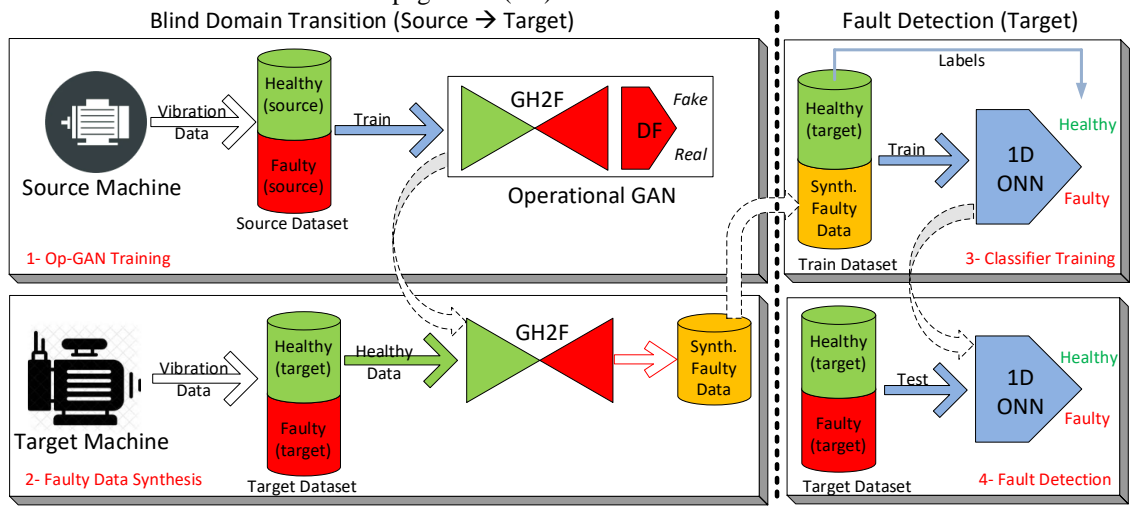

**Figure 2:** The illustration of the proposed zero-shot fault detection by blind domain transition.

The general framework of the proposed blind domain transition and fault detection scheme is illustrated in Figure 2. The proposed approach follows a segment-based transition and fault detection scheme where each segment has a one-second duration without overlaps. With the sampling frequency of 4096 Hz, this corresponds to $m = 4096$ samples per segment. Using the labeled segments on the source machine(s), in the first stage, 1D Op-GAN is trained to form a generator and discriminator. With proper training, the generator learns the natural transition from any healthy signal to the corresponding faulty signal under various working conditions, sensor parameters, and faulty types/severities. In the second stage, this generator can then be used to synthesize the *potential* faulty signals from the corresponding healthy signals of the target (unseen) machine. As shown in Figure 2, the healthy and faulty vibration data are first partitioned into 1-second segments. In order to accomplish the magnitude invariance due to the different sensor locations, each segment is first linearly normalized as follows:

$$X_N(i) = \frac{2(X(i) - X_{\min})}{X_{max} - X_{min}} - 1 \qquad (8)$$

where $X(i)$ is the $i^{th}$ original sample amplitude in the segment, $X_N(i)$ is the $i^{th}$ sample amplitude of the normalized segment, $X_{min}$ and $X_{max}$ are the minimum and maximum amplitudes within the segment, respectively. This will scale the segment linearly in the range of [-1 1] where $X_{min} \rightarrow -1$ and $X_{max} \rightarrow 1$. The proposed Op-GAN model consists of two Self-ONNs: the generator and the discriminator. As mentioned earlier, while the generator learns how to map healthy motor signals to faulty counterparts while preserving the characteristics of the signal, the discriminator aims to maximize the loss functions to generate more realistic transformations. The objective function of the Op-GAN can be expressed as,

$$\min_G \max_D \gamma_{cGAN}(G,D)\\ = E[\log(D(X,Y))]\\ + E[\log(1 - D(X, G(X,z)))] \qquad (9)$$

where $G$ is the generator, $D$ is the discriminator, $z$ is the 1D random noise and $Y$ is the corresponding output label. While the generator is aiming to minimize the loss term, $\log(1 - D(X, G(X,z)))$, the discriminator's target is to maximize both terms of $D(X,Y)$ and $\log(1 - D(X, G(X,z)))$.

$$STFT[X,w,n] = X(n,w) = \sum_m X[m]W[n-m]e^{-jwm} \qquad (10)$$

$$X(n,k) = X(n,w)|_{w=\frac{2\pi k}{N}} \rightarrow Spec(X(n,k)) = |X(n,k)|^2 \qquad (11)$$

$$Loss_{Time} = \|(Y(z) - G(X,z))\|_1 \qquad (12)$$

$$Loss_{STFT} = \|STFT(Y(z)) - STFT(G(X,z))\|_1 \qquad (13)$$

To generate and then detect the faulty signals better, both temporal and spectral signal representations are used. In this approach, it is important to note that the generator aims to minimize the MAE error both in time (by Eq. (12)) and in frequency (by Eq. (13)) domains. As for spectral representation, we first take *N*-points discrete Short-time Fourier Transform (STFT) of both the input and output signals. Eq. (10) formulates the continuous STFT where the $X$ is the input signal, $W$ is the window function. Consequently, Eq. (11) expresses the *N*-point discrete STFT which is complex-valued. We, therefore, used the spectrogram, $Spec(X(n,k))$, for the final spectral representation. As for the window function, we used the *N*=256 samples long *Hanning* window with an overlap of 128 samples, and thus, *N*-point discrete Spectrogram is computed. Finally, the overall loss function used in the Op-GAN training is expressed as follows:

$$Loss_{total} = Loss_{BCE} + \lambda (Loss_{Time} + Loss_{STFT}) \qquad (14)$$

where $\lambda$, is the weight parameter which balances the temporal and spectral loss functions with the BCE loss.

*C. Zero-shot Fault Detection*

As a result of such blind domain transition, the dataset needed for training the fault detector can then be formed using *real* healthy signals and *synthetic* (generated) faulty signals. As illustrated in the right side of Figure 2, during the third stage, over the train dataset a 5-operational-layer 1D Self-ONN model is trained directly on the *target* domain as the fault detector. As mentioned earlier, due to the elegant domain transition performance of the proposed approach, the compact Self-ONN can then be used to detect any (actual) faulty signal during the final (4[th]) stage, which is the core operation of the continuous health monitoring on the target machine. The detector model features with train and hyper-parameters will be presented in Section III.

Finally, for a new (target) machine, it is sufficient to repeat only the 2[nd] and 3[rd] stages to train the dedicated classifier for fault detection on that machine. This further yields an elegant computational efficiency since the Op-GAN training can now be skipped and only another compact 1D Self-ONN model should be trained, which can be implemented in real-time without requiring any specific hardware. The computational complexity analysis of the Op-GAN and detector model will be presented in Section III.E.

## III. EXPERIMENTAL RESULTS

In this section, the benchmark motor bearing fault dataset used in this study will be introduced first. Then, the experimental setup used for the evaluation of the proposed bearing fault severity classification framework will be presented. The comparative evaluations and the overall results of the experiments obtained using the real motor vibration signals from the QU-DMBF dataset will be presented in Section III.C. Finally, the computational complexity of the proposed approach for both training (offline) and classification (online) phases will be evaluated in detail in Section III.D.

*A. Qatar University Dual-Machine Bearing Fault Benchmark Dataset: QU-DMBF*

The benchmark dataset used in this study was generated by the researchers at Qatar University from two electric machines. Figure 3 shows the installation of two machines (A and B) along with the sensor locations. For Machine-A, the setup includes a 3-phase AC motor (brand: VEMAT, Model: 3VTB-90LA(2P), Vicena-Italy), whose input frequency was controlled by a variable frequency drive. The motor used in this experiment supplied 2.2 kW (3 Hp) at a maximum speed of 2840 rpm. The vibration signals were acquired by using 5 high sensitivity, ceramic shear ICP accelerometers (from PCB Piezotronics, Model No. 352C33, 100 mV/g, NY-USA), which were fixed on the same mounting base that supported the load cell (from Omegadyne Inc, Model No. LCM204-50KN, Manchester-UK), and 4 other different location on machine and motor. The readings were controlled by two four-channel NI-9234 sound and vibration input modules at a sampling frequency of 4.096 kHz. It has a ~180 kg weight and the setup has 100x100x40 cm dimensions. Machine-A has the following variations on the working conditions:

- 19 different bearing configurations: 1 healthy, 18 fault cases: 9 with a defect on the outer ring, and 9 with a defect on the inner ring. The defect sizes are varying from 0.35mm to 2.35mm.
- 5 different accelerometer localization: 3 different positions and 2 different directions (radial and axial)
- 2 different load (force) levels: 0.12 kN and 0.20 kN.
- 3 different speeds: 480, 680, and 1010 RPM.

For each working condition of a healthy bearing, we acquire data for 270 seconds to form a *healthy* record. Therefore, this yields a total duration of 270 x 5 x 2 x 3 = 8,100 seconds of healthy vibration data and 30 *healthy* records, each per working condition. For each working condition of a faulty bearing, we acquire data for 30 seconds. Therefore, this yields a total duration of 30 x 18 x 5 x 2 x 3 = 16,200 seconds of *faulty* data, and 540 *faulty* records, each per working condition. Therefore, faulty data has twice more duration of healthy data. As a result, the total duration of the dataset for Machine-A is 24,300 seconds (6.75 hours).

Machine B, on the other hand, was originally a Machinery Fault Simulator from SpectraQuest Inc, USA. The setup includes a DC motor of 0.37 kW (0.5 HP), 90 VDC, 5 A, with a maximum rotational speed of 2500 RPM. The original shaft of the machine was removed and replaced by a new and bigger one to accommodate the same bearings used on Machine-A. For the sake of resistance and stability, other supporting mechanical components were also redesigned. The machine has an approximate weight of 50kg and overall dimensions of 100x63x53 cm. Machine-B has the following variations on the working conditions:

- 19 different bearing configurations: 1 healthy, 9 with a defect on the outer ring, and 9 with a defect on the inner ring. The defect sizes are varying from 0.35mm to 2.35mm.
- 6 different accelerometer positions.
- A fixed load (force) of 0.15 kN.
- 5 different speeds: 240, 360, 480, 700, and 1020 RPM.

For each working condition for a healthy bearing, we acquire data for 270 seconds to form a *healthy* record. Therefore, this yields a total duration of 270 x 6 x 1 x 5 = 8,100 seconds of *healthy* vibration data and 30 healthy records. For each working condition for each faulty bearing, we acquire data for 30 seconds to form a *faulty* record. Therefore, this yields a total duration of 30 x 18 x 6 x 1 x 5 = 16,200 seconds of *faulty* data, and 540 *faulty* records - once again making a 2:1 ratio over the healthy data. As a result, the total duration of the dataset for machine B is 24,300 seconds (6.75 hours).

Besides having the data of the two entirely different electric machines, the QU-DMBF dataset has unique features compared to the aforementioned benchmark datasets. First, it has a record number of variations in working conditions, e.g., 540 different conditions only for the bearing faults in both machines. Sensor locations and numbers also show high variations as shown in Figure 3. This is the reason for the significantly varying temporal and spectral patterns of the vibration data not only between the two machines but also within the same machine (e.g., see Figure 1 for some samples). This feature especially would make any conventional approach fail when the classifier is not trained on the particular condition for which it is tested. Finally, this dataset has one of the longest acquisition duration ever recorded with a total of 13.5 hours. The QU-DMBF is now publicly shared in [57] to serve as the dual-machine bearing fault detection benchmark for the research community.

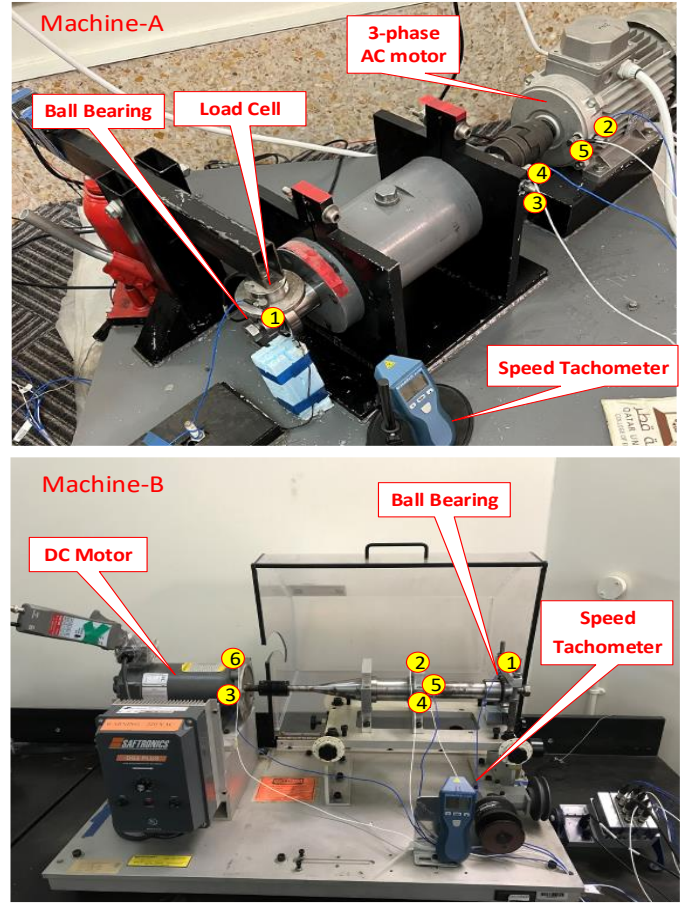

**Figure 3: Machines A and B where the locations of the sensors are highlighted with circles.**

### B. Experimental Setup

For the 1D Op-GAN, the generator has the 10-layer U-Net configuration with 5 1-D operational layers and 5 transposed operational layers with skip connections. The kernel sizes are set as 5 except that in the last transposed layer the kernel size 6 is used. The stride is set as 2 for both operational and transposed operational layers. The kernel sizes of the discriminator are set as 4 except that in the last transposed layer the kernel size 6 is used. The stride for layers is set as 4, 4, 4, 4, 4, and 2 respectively. The architectures for the generators and discriminators are shown in Figure 4. For all experiments, we employ a training scheme with a maximum of 1000 BP iterations with batch size 8. Adam optimizer with an initial learning rate of $10^{-4}$ is used for training both generator and discriminator via Back-Propagation (BP) (in stage 1 in Figure 2). The parameter $\lambda$ in Eq. (14) is set to 100. We implemented the proposed 1D Self ONN architectures using the FastONN library [47] based on PyTorch. As the training data for the Op-GAN, 5400 healthy and faulty segments with a duration of one second (4096 samples) are selected in the source machine. Within the train set, all records belonging to one of the speed settings are spared as the validation set. The 1D Self-ONN with only 5 operational layers and 2 dense layers is used as the fault detector (see Figure 5). It has 16 neurons in each hidden operational layer, and 32 neurons in the hidden dense layer. Meanwhile, the output layer has two neurons for this binary classification problem. The neuron in the input layer takes a 1-second vibration segment. The nonlinear activation function "*tanh*" is used in all layers. The kernel sizes of the operational layers are set as 81, 41, 21 7, and 7 respectively. The strides of operational layers are set as 8, 4, 4, 2, and 2, respectively. For training the detector on the target machine (in stage 3 in Figure 2), Adam optimizer is used as the optimizer for Back-Propagation (BP) with the initial learning factor, $\varepsilon$, set to $10^{-4}$. Mean-Squared-Error (MSE) is used as the loss function and the maximum number of BP iterations (epochs) is set to 50.

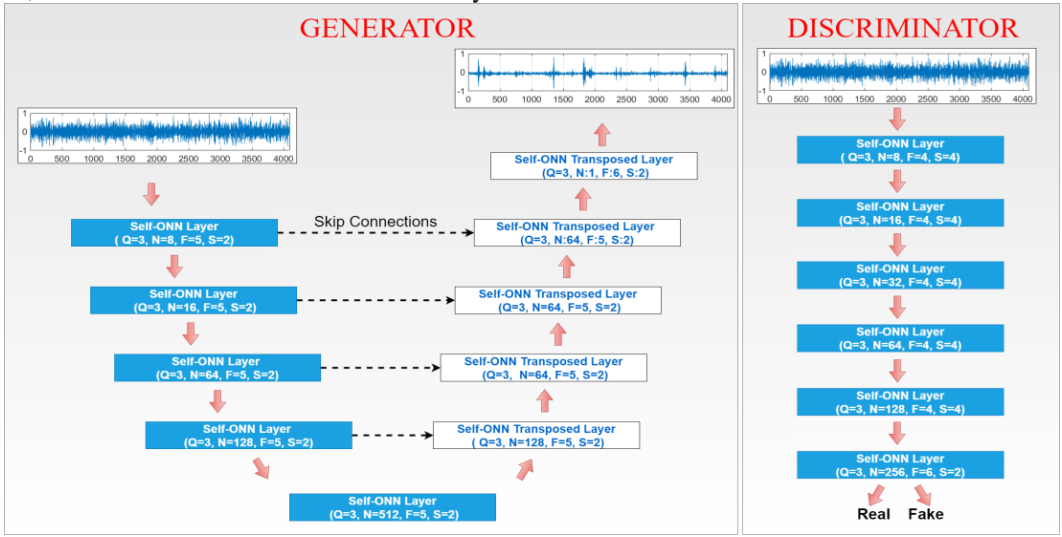

**Figure 4:** The Generator and Discriminator architectures of the Op-GAN.

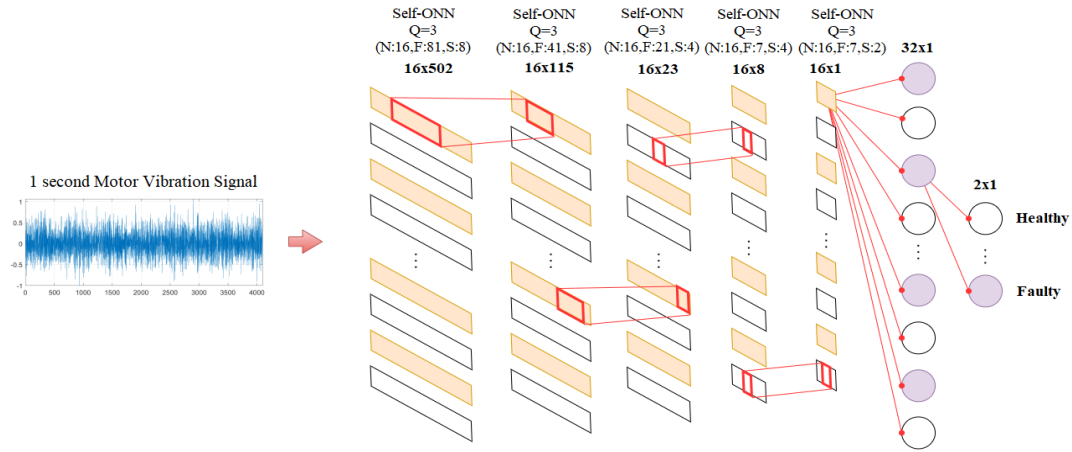

**Figure 5:** The configuration of the 1D Self-ONN fault detector.

## C. Domain Transition Results

In both blind domain transitions, once the Op-GAN is trained on the source machine, its generator is used to synthesize the artificial fault signals of the target machine from its *real* healthy signals as illustrated in Figure 2. We used 71 % of the healthy signals randomly selected from all 30 healthy cases to synthesize the fault signals (in stage 2 in Figure 2) and the rest is used for testing the fault detection performance on the target machine (in stage 4 in Figure 2). For a high zero-shot fault detection performance, it is crucial that the generator should be able to synthesize such fault signals that share similar temporal and spectral patterns with the (actual) fault signals from the corresponding healthy signal of that working condition.

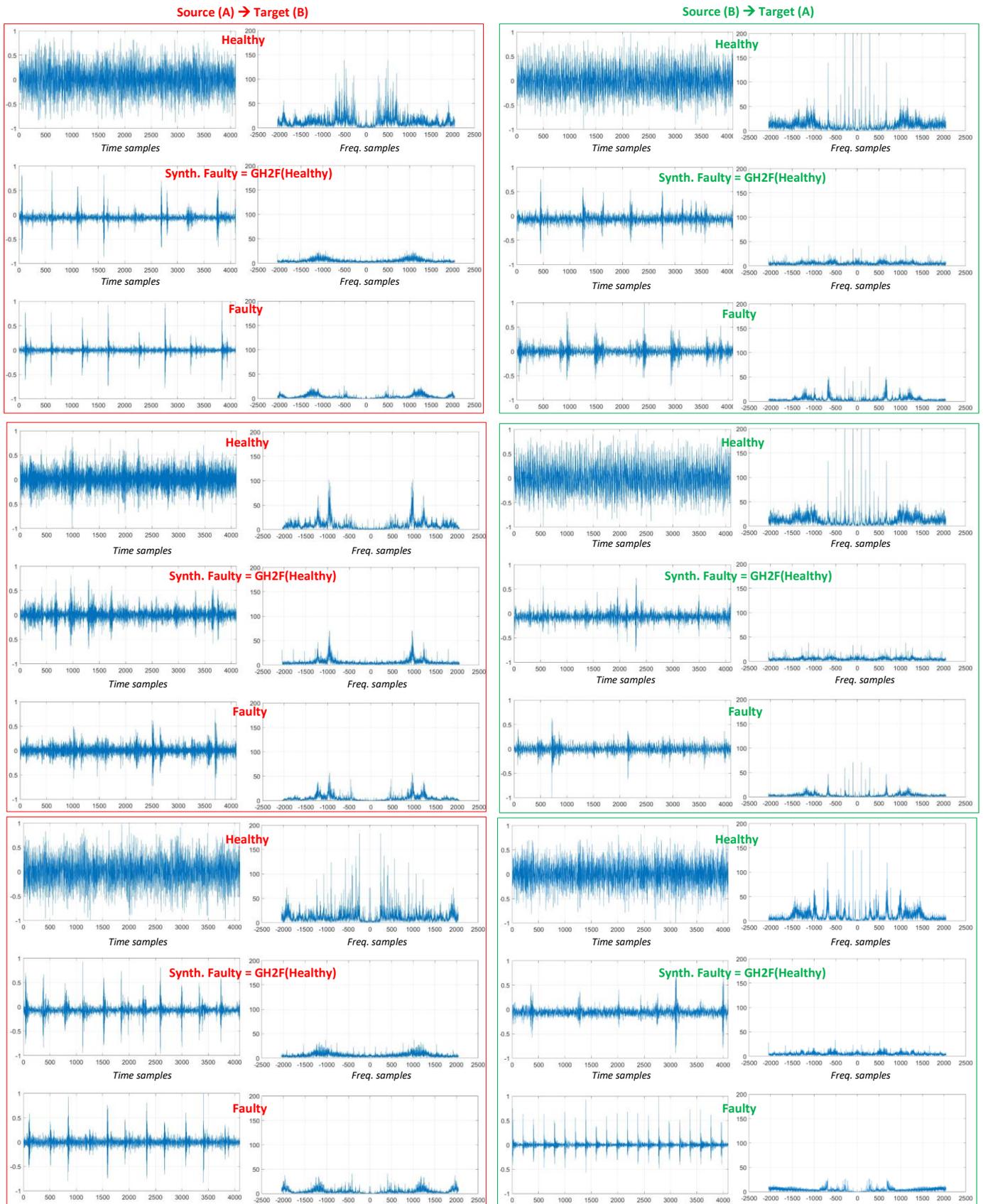

**Figure 6: 6 typical samples (three on the left column: target = B, three on the right column: target = A) of the fault signal synthesis. Each sample has a plot from the healthy signal (top) and the corresponding synthesized (middle) and actual (bottom) fault signals both in time (left plots) and frequency (right plots) domains.**

Figure 6 shows six typical samples (three from each target machine) of the fault signal synthesis from the healthy signals and the corresponding (actual) fault signals in both time and frequency domains. The three samples on the left column are from the target machine-B and taken from sensors 1, 3, and 5. The first two faults belong to the inner ring and the one on the bottom belongs to the outer ring. The other three samples on the right column are from the target machine-A and taken from sensor 1 but with two different speeds. The first two faults belong to the inner ring and the one on the bottom belongs to the outer ring. Even though all healthy and faulty conditions are from entirely different conditions, the samples clearly demonstrate that the Op-GAN trained on the source machine can indeed synthesize such fault signals on the target machine that are similar to their actual counterparts. This further reveals the fact that the Op-GAN can successfully capture the core characteristics of the natural transition between healthy and the corresponding faulty states in both source machines. Thus, it can accurately generate any fault state of an unseen (target) machine using only its (corresponding) healthy signal.

### D. Fault Detection Results

The evaluation of the proposed zero-shot fault detection method is performed over both machines by switching the source and target machines accordingly. So, in the first set of experiments, Machine A is used as the source domain where stages 1-3 are performed, and Machine-B is used as the target domain where stage-4 (testing) is carried out using the compact 1D Self-ONN fault detector. The machine assignments are simply swapped for the second set of experiments. For each set, we consider the fault detection per working condition, i.e., within the 30-seconds record. An important fact in fault detection is that once the bearing becomes "faulty," it will not recover to a "healthy" status since the damaged bearings can only get worse in time. Therefore, once the faulty condition is detected within any 30-second record, even for only one segment, the entire record can be classified as *faulty* and other segments that are misclassified as "healthy" can then be ignored. To minimize occasional false alarms, we classify each record as *faulty* if more than one segment is classified as faulty; otherwise, the record is classified as *healthy*. Therefore, we can report the number of faulty conditions (records) that are classified accurately among all 540 faulty conditions. This will give us the record-wise *Recall* rate of fault detection. However, since we used around 71% of the target motor's healthy data to train the fault detector, we will report the segment-based detection performance over the healthy segments in the test set. Moreover, from the number of misclassified healthy segments in the test set (with 6000 records) we compute the false alarm rate (*FAR*).

When the Op-GAN is trained over the Machine-A (source), its generator model at epoch 430 yielded the best detection performance over the validation set, and therefore, was used for fault synthesis for Machine-B (target). Table 1 presents the number of accurately detected fault records per sensor over Machine-B. As there are 6 sensors (accelerometers) on the machine, each sensor has acquired 90 faulty conditions. Figure 3 (top) shows the sensor locations. The corresponding *Recall* rates per sensor are plotted in Figure 7. No healthy frames were misclassified as faulty. Hence, the average fault detection *Recall* and *FAR* can be computed as 95.1% and 0%, respectively for Machine-B. The average fault detection *Precision* for Machine-B is, therefore, 100%.

**Table 1: No. of accurately detected fault records per sensor in the (target) Machine-B.**

| Acc. 1 (90) | Acc. 2 (90) | Acc. 3 (90) | Acc. 4 (90) | Acc. 5 (90) | Acc. 6 (90) | Total (540) |
|---|---|---|---|---|---|---|
| 87 | 80 | 87 | 81 | 89 | 90 | **514** |

**Table 2: No. of accurately detected fault records per sensor in the (target) Machine-A.**

| Acc. 1 (108) | Acc. 2 (108) | Acc. 3 (108) | Acc. 4 (108) | Acc. 5 (108) | Total (540) |
|---|---|---|---|---|---|
| 108 | 59 | 81 | 99 | 62 | **409** |

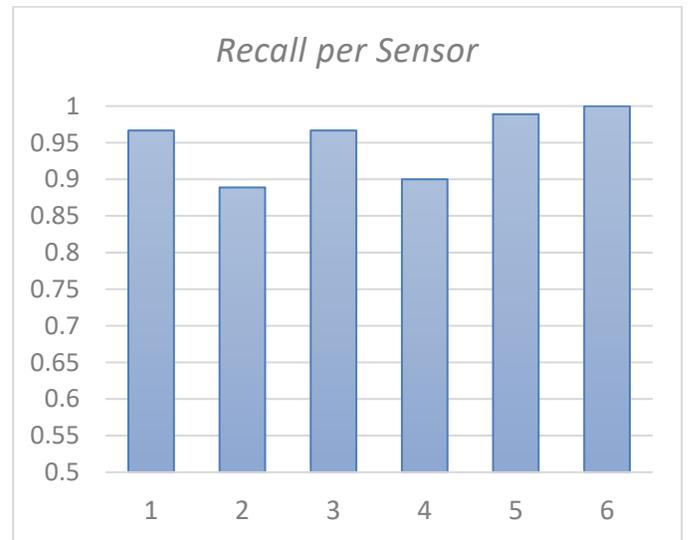

Figure 7: *Recall* rates of fault detection for each sensor on the (target) Machine-B.

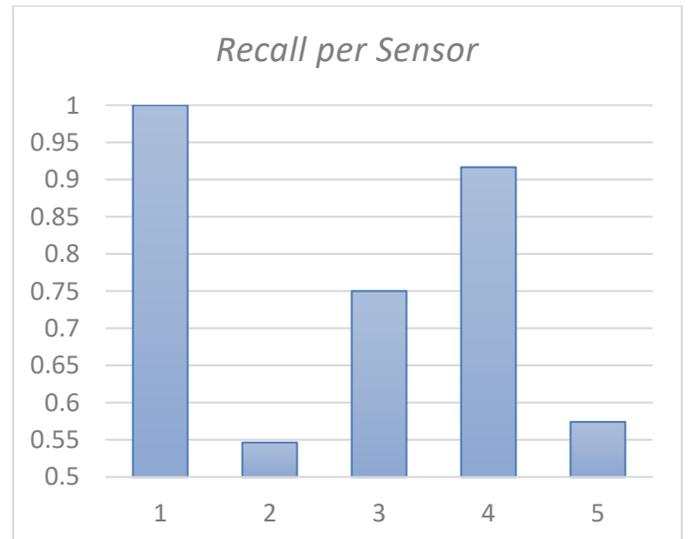

Figure 8: *Recall* rates of fault detection for each sensor on the (target) Machine-A.

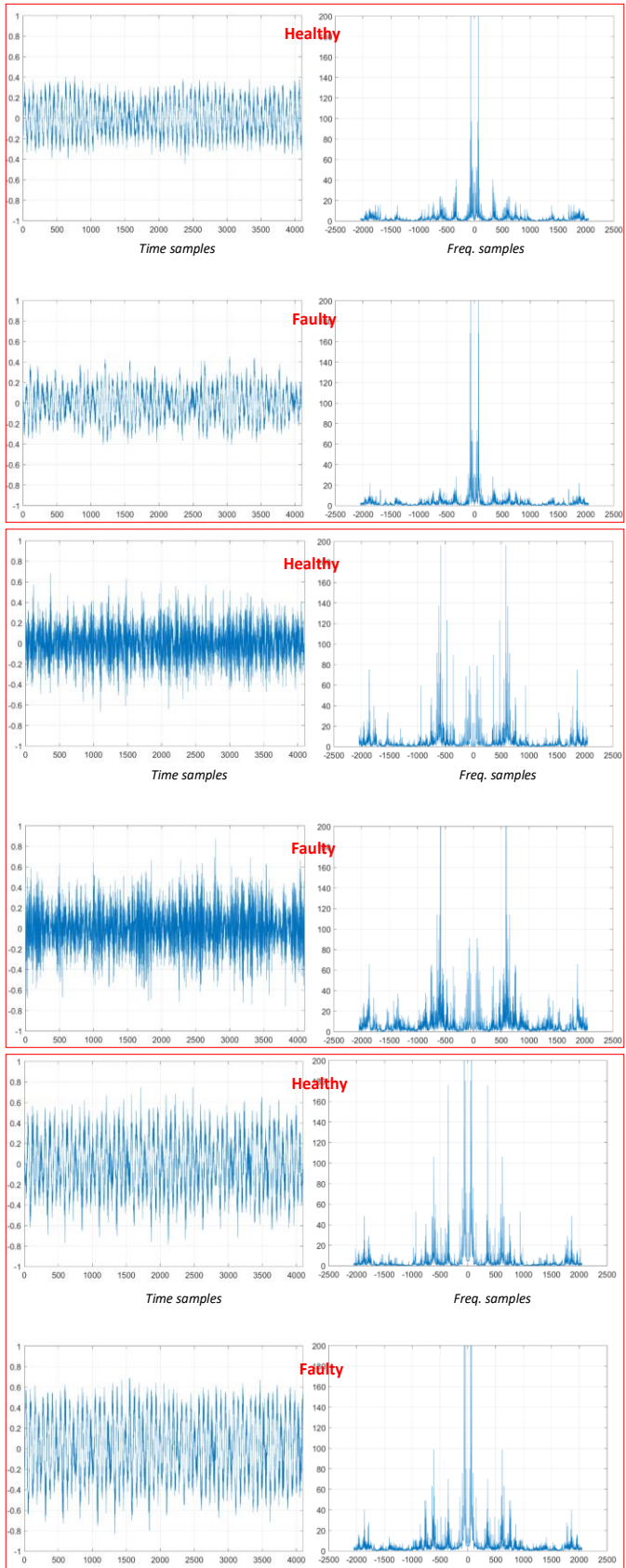

**Figure 9:** Acquired from sensors 2 and 5 in Machine-A, three typical samples of the healthy and faulty signals each from the same working condition.

When the Op-GAN is trained over Machine-B (source), its generator model at epoch 630 yielded the best detection performance over the validation set and hence, was used for fault synthesis for Machine-A (target). Table 2 presents the number of accurately detected fault records per sensor over Machine-A. As there are 5 sensors (accelerometers) on this machine, each sensor has acquired 108 faulty conditions. Figure 3 (bottom) shows the sensor locations. The corresponding *Recall* rates per sensor are plotted in Figure 8. Only 38 segments were misclassified among 6000 healthy segments in the test set. Hence, the overall fault detection *Recall* and *FAR* can be computed as 75.7% and 0.6%, respectively for Machine A. The average fault detection *Precision* for Machine-A is, therefore, 99.6%.

Over the fault detection results, various important observations can be made. First of all, achieving above 95% recall rate on fault detection with no false alarms shows that the domain transition learned over the (source) Machine-A indeed captures almost all transition characteristics of the (target) Machine-B. However, when the domain transition is learned over the Machine-B, the fault detection recall is dropped to 75.7% with an insignificant level of false alarms. The major reason behind this drop is the location of the sensors 2 and 5, both of which are on the motor. Hence the corresponding recall rates from both sensors are below 60% as shown in Figure 8. Considering the significantly heavier bulk of Machine-A (around 150kg), this is an expected outcome because being on top of a heavy motor and at the farthest point from the ball bearing, those two sensors will naturally be insensitive especially to the minor faults (faults with small diameters) on the bearings. The three typical samples (two from sensor 2 and one from sensor 5 in Machine A) of the healthy and faulty signals, each from the same working condition (top and bottom plots are from inner, the middle plot is from the outer ring, respectively) are shown in Figure 9. Both healthy and faulty waveforms are almost identical, and it is, therefore, not possible for a zero-shot detector to discriminate such a fault signal from the healthy. This is true even for a local detector trained directly over the actual faulty signals of the Machine-A. To accommodate this, we train the same detector model using the same hyper- and train parameters over the identical healthy/faulty partitions of the Machine-A, and it can reach up to an average 93.65% recall, once again failing to discriminate mostly on the vibration data from sensors 2 and 5. Note further that 93.65% recall rate can be considered as the upper bound for the zero-shot fault detection, it has the advantage of being trained over the *actual* fault data. As a result, this shows the importance of the sensor locations especially for heavy machinery. Without those two sensors, the proposed zero-shot fault detection achieves an average recall of 88.89%.

*E. Computational Complexity Analysis*

For computational complexity analysis, the network size, total number of parameters (PARs), and inference time for each network configuration are computed and reported the detailed formulations of the PARs calculations for Self-ONNs can be found in [46]. All the experiments were carried out on a 2.2 GHz Intel Core i7 with 16 GB of RAM and NVIDIA GeForce RTX 3080 graphic cards. The Op-GAN network is implemented using Pytorch and FastONN library. The generator and the discriminator network of the GAN consist of 244K and 133K parameters respectively. However, since the discriminator will be discarded afterward, discriminator network parameters can be omitted. The fault detector network contains around 63K parameters. For a single CPU implementation, specifically, the required amount of time to synthesize a 1-second *faulty* segment using the generator and classify it using the detector takes about 0.7 msec and 0.089 msec, respectively. This indicates more than 1400 and 11000 times the real-time speed on the test computer and hence, both of the proposed (blind) domain transition and (zero-shot) fault detection have, therefore, the potential to be implemented on mobile, low-power devices in real-time. Especially, the fault detector can be a software plugin for any sensor, thus performing not only the acquisition but also the fault detection in real-time. Such smart sensors can build the backbone of continuous motor health monitoring for any (new) machine just after it becomes operational.

## IV. CONCLUSIONS

Despite numerous fault detection methods being proposed in the literature, they are only applicable to a single (target) machine on which they were trained in advance. Their detection performance usually degrades significantly even when one or a few of the conditions are altered. However, what makes them impractical is the assumption of having the labeled fault data in advance for any machine. Recent DA approaches attempted to address this problem by transferring knowledge from the source domain (machine) data where the labeled fault data is available to the target domain (machine) data drawn from a different but assuming *a related distribution*. Ironically, most of them were still evaluated on the same machine (dataset) while altering usually a single working condition at a time to maintain the assumption of *the related distribution,* but for some cases, drastically low fault detection accuracies were still encountered. Finally, regardless of the DA method used, the performance significantly got worse when compact/shallow classifiers are used for detection.

This study has followed a completely different path from the prior DA works, which basically adapt or transform the target features to the source domain so that the source classifier can be used for detection. Instead, the knowledge (the transition from the healthy to faulty signal) learned in the source machine is transferred directly to the target machine to synthesize its potential faulty signals for any working condition. Accomplishing such a blind domain transition using 1D Op-GANs thus allows us to train the classifier directly on the target domain. This eliminates the need for any feature extraction and adaptation employing certain transformations, discriminations, and possibly other tricks such as data augmentation. Moreover, due to the elegant performance of the proposed blind domain transition, deep and complex classifiers are no longer needed and thus, a compact and lightweight 1D Self-ONN model can conveniently be used to detect any *actual* bearing fault in real-time. To validate our objectives, we create one of the largest bearing fault datasets encapsulating vibration data of the two different machines, acquired under 540 distinct working conditions, i.e., different sensor locations, speed, load, and fault severities. We publicly shared it with the research community in [57] to serve as the benchmark dataset. The experimental results have demonstrated that our objectives have been achieved with high precision-recall rates and elegant computational efficiency. In particular, the results show that the locations of the sensors play an important role for an accurate detection, especially on a heavy machine. So, the detection performance can significantly be improved when the sensors are placed on or in a close proximity to the parts targeted for fault detection (i.e., bearings in this study).

While the proposed approach has achieved around 95% and 76% (or ~89% without the two problematic sensors) average recall rates with no or insignificant false alarms, the performance can still be improved if the domain transition would be learned over several (source) machines. In this way, it will be probable to learn those *missed* healthy-faulty transition characteristics required by the target machine, thus leading to a better fault synthesis and a higher fault detection performance. Although this study has focused on bearing fault detection over the vibration signals, the proposed approach can conveniently be used for other signals (e.g., acoustics, current, etc.), fault types, and other related applications, e.g., for *structural* health monitoring where the fault/anomaly data is not available or very difficult to obtain for a particular structure. A natural extension will be to perform fault/anomaly *recognition* to further identify the type and predict the severity of the fault and its location for complete real-time health monitoring. These will be the topics of our future research.

IEEE TRANSACTIONS ON INDUSTRIAL ELECTRONICS 13[52] M. Uzair, S. Kiranyaz and M. Gabbouj, "Global ECG Classification by Self-Operational Neural Networks with Feature Injection", eprint arXiv:2204.03768, April 2022.

[53] M Soltanian, J Malik, J Raitoharju, A Iosifidis, S Kiranyaz, M Gabbouj, "Speech Command Recognition in Computationally Constrained Environments with a Quadratic Self-Organized Operational Layer", in Proc. of Int. Joint Conference on Neural Networks (IJCNN), pp. 1-6, July 2021.

[54] X. Jiang, D. Wang, D. T. Tran, S. Kiranyaz, M. Gabbouj and X. Feng, "Generalized Operational Classifiers for Material Identification," 2020 IEEE 22nd Int. Workshop on Multimedia Signal Processing (MMSP), pp. 1-5, Tampere, 2020. doi: 10.1109/MMSP48831.2020.9287058.

[55] J. Malik, S. Kiranyaz, M. Gabbouj, "BM3D vs 2-Layer ONN", in Proc. of IEEE Int. Conference on Image Processing (ICIP), Sep. 2021. DOI:10.1109/ICIP42928.2021.9506240

[56] S. Kiranyaz, O. C. Devecioglu, T. Ince, J. Malik, M. Chowdhury, T. Hamid, R. Mazhar, A. Khandakar, A. Tahir, T. Rahman, and M. Gabbouj, "Blind ECG Restoration by Operational Cycle-GANs", in arXiv Preprint, https://arxiv.org/abs/2202.00589 , Feb. 2022.

[57] Paszke, A., Gross, S., Massa, F., Lerer, A., Bradbury, J., Chanan, G., Killeen, T., Lin, Z., Gimelshein, N., Antiga, L., Desmaison, A., Köpf, A., Yang, E., DeVito, Z., Raison, M., Tejani, A., Chilamkurthy, S., Steiner, B., Fang, L., Bai, J., Chintala, S., 2019. PyTorch: An imperative style, high-performance deep learning library. arXiv.

[58] Zero Shot Bearing Fault Detection by Blind Domain Transition, Version 1.0, Source code. [Online]. Available: https://github.com/OzerCanDevecioglu/Zero-Shot-Bearing-Fault-Detection-by-Blind-Domain-Transition

[59] S. Kiranyaz, T. Ince, A. Iosifidis and M. Gabbouj, "Operational Neural Networks", Neural Computing and Applications (Springer-Nature), Mar. 2020. DOI: https://doi.org/10.1007/s00521-020-04780-3

[60] S. Kiranyaz, J. Malik, H. B. Abdallah, T. Ince, A. Iosifidis, M. Gabbouj, "Exploiting Heterogeneity in Operational Neural Networks by Synaptic Plasticity", Neural Computing and Applications (Springer-Nature), pp. 1-19, Jan. 2021. https://doi.org/10.1007/s00521-020-05543-w

[61] J. Malik, S. Kiranyaz, M. Gabbouj, "Operational vs Convolutional Neural Networks for Image Denoising", arXiv:2009.00612, Sep. 2020.

[62] S. Kiranyaz, T. Ince, A. Iosifidis and M. Gabbouj, "Generalized model of biological neural networks: progressive operational perceptrons," IJCNN, 2017

[63] S. Kiranyaz, T. Ince, A. Iosifidis and M. Gabbouj, "Progressive operational perceptrons," *Neurocomputing*, vol. 224, pp.142–154, 2017.

[64] D.T. Tran, S. Kiranyaz, M. Gabbouj and A. Iosifidis, "Progressive operational perceptron with memory," *Neurocomputing* 379, pp. 172-181, 2020.

[65] D.T. Tran and A. Iosifidis, "Learning to Rank: A progressive neural network learning approach", IEEE Int. Conf. on Acoustics, Speech, and Signal Processing, Brighton, U.K., 2019.

[66] D.T. Tran, S. Kiranyaz, M. Gabbouj and A. Iosifidis, "Heterogeneous Multilayer Generalized Operational Perceptron", *IEEE Trans. on Neural Networks and Learning Systems*, pp. 1-15, May 2019. DOI: 10.1109/TNNLS.2019.2914082.

[67] D.T. Tran, S. Kiranyaz, M. Gabbouj and A. Iosifidis, "Knowledge Transfer For Face Verification Using Heterogeneous Generalized Operational Perceptrons", IEEE Int. Conf. on Image Processing, Taipei, Taiwan, 2019.

[68] S. Kiranyaz, T. Ince, and M. Gabbouj, "Personalized Monitoring and Advance Warning System for Cardiac Arrhythmias," Scientific Reports - Nature, vol. 7, Aug. 2017. DOI 10.1038/s41598-017-09544-z (SREP-16-52549-T) (http://rdcu.be/vfYE).